\documentclass{article} % For LaTeX2e
\usepackage{graphicx}
\usepackage{subcaption}
\usepackage{xcolor}
\usepackage{comment}
\usepackage{booktabs}
\usepackage{wrapfig}
\usepackage{multirow}
\usepackage{iclr2024_conference,times}
\iclrfinalcopy

\usepackage{amsmath,amsfonts,bm}

\def\eqref#1{equation~\ref{#1}}
\def\1{\bm{1}}

\DeclareMathAlphabet{\mathsfit}{\encodingdefault}{\sfdefault}{m}{sl}
\SetMathAlphabet{\mathsfit}{bold}{\encodingdefault}{\sfdefault}{bx}{n}

\usepackage{hyperref}
\usepackage{url}

\title{Chain of Images for Intuitively Reasoning}

\author{Fanxu Meng$^{1,2}$, Haotong Yang$^{1}$, Yiding Wang$^{1}$, Muhan Zhang$^{1,2}$\\
$^{1}$School of Intelligence Science and Technology, Peking University\\
$^{2}$National Key Laboratory of General Artificial Intelligence, BIGAI\\
\centering\href{https://github.com/GraphPKU/CoI}{https://github.com/GraphPKU/CoI}}
\begin{document}

\maketitle

\begin{abstract}
The human brain is naturally equipped to comprehend and interpret visual information rapidly.
When confronted with complex problems or concepts, we use flowcharts, sketches, and diagrams to aid our thought process.
Leveraging this inherent ability can significantly enhance logical reasoning.
However, current Large Language Models (LLMs) do not utilize such visual intuition to help their thinking. Even the most advanced version language models (e.g., GPT-4V and LLaVA) merely align images into textual space, which means their reasoning processes remain purely verbal.
To mitigate such limitations, we present a Chain of Images (CoI) approach, which can convert complex language reasoning problems to simple pattern recognition by generating a series of images as intermediate representations.
Furthermore, we have developed a CoI evaluation dataset encompassing 15 distinct domains where images can intuitively aid problem-solving. Based on this dataset, we aim to construct a benchmark to assess the capability of future multimodal large-scale models to leverage images for reasoning.
In supporting our CoI reasoning, we introduce a symbolic multimodal large language model (SyMLLM) that generates images strictly based on language instructions and accepts both text and image as input.
Experiments on Geometry, Chess and Common Sense tasks sourced from the CoI evaluation dataset show that CoI improves performance significantly over the pure-language Chain of Thoughts (CoT) baselines.
\end{abstract}
\begin{figure}[hbt]
    \centering
    \includegraphics[width=0.9\textwidth]{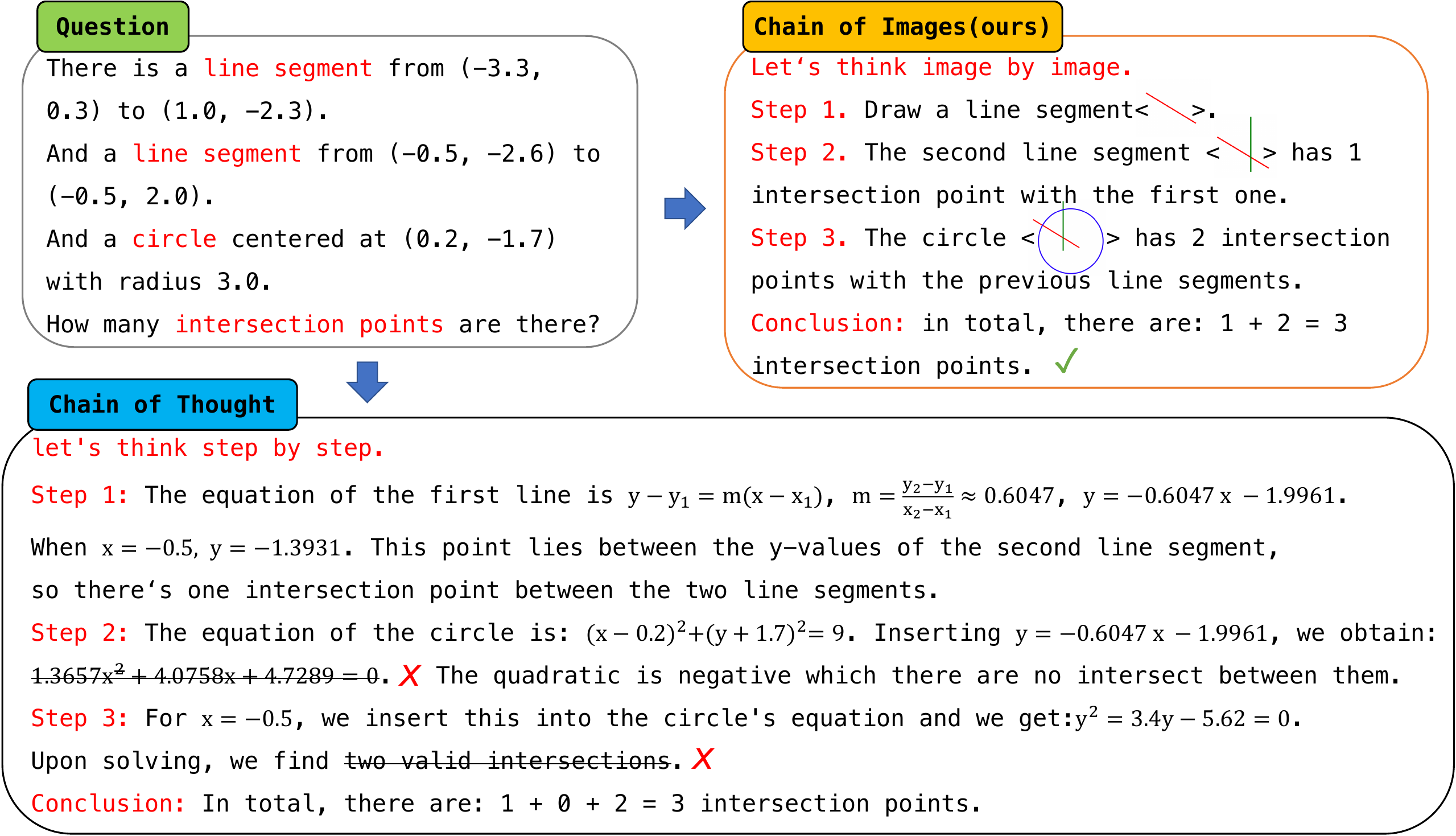}
    \caption{The simplified steps employing GPT-4 to calculate the number of intersection points using CoT are displayed at the bottom of the image.
    The whole processes of CoI are displayed at the right-top of the image.
    Two issues were identified in CoT:
    1. Incorrect numerical values were used during the formula derivation, and
    2. The existence of endpoints in the line segments was overlooked.
    By contrast, CoI easily identifies the number of intersection points in the image generated by SyMLLM.}
    \label{fig:cot-coi}
\end{figure}

\section{Introduction}
Large-scale pre-trained language models (LLMs)~\citep{vaswani2017attention,devlin2018bert,brown2020language,chowdhery2022palm,touvron2023llama,openai2023gpt4} have found applications in diverse areas. 
Beyond their robust foundational capabilities~\citep{kaplan2020scaling,ouyang2022training}, a significant advancement lies in the evolution of the prompting technique~\citep{liu2023pre}, which provides LLMs with a context to guide their responses and reasoning processes. 
As presented by~\citet{brown2020language}, the few-shot prompting approach offers the model several question-answer pairs to guide its outputs. 
The Chain-of-Thought (CoT) prompting strategy~\citep{wei2022chain} provides sequential, step-by-step answer examples, enabling the model to perform intricate multi-step reasoning.
The zero-shot CoT technique~\citep{kojima2022large} omits to provide examples, instead stimulating the model to think progressively with a simple ``Let's think step by step'' prompt.
For instance, in Figure~\ref{fig:cot-coi}, to accurately count the intersection points of geometric shapes, CoT follows a sequential approach: 
1) Determine the formula of each shape, 
2) Compute the intersection points between each pair of shapes, and 
3) Sum up all the intersection points.

\begin{figure}[hbt]
\begin{minipage}{0.5\textwidth}
    \includegraphics[width=0.95\linewidth]{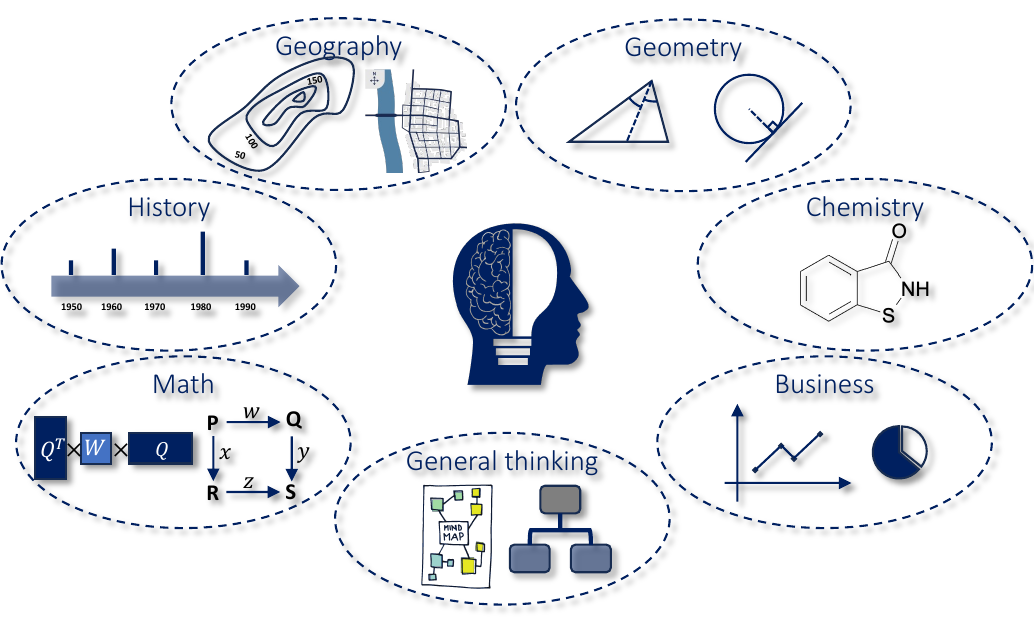}
    \caption{Images play a pivotal role in many disciplines. We tend to imagine pictures to solve problems intuitively.}
    \label{fig:helpful_image}
\end{minipage}
    \hspace{0.01\linewidth} 
\begin{minipage}{0.5\textwidth}
    \includegraphics[width=0.95\linewidth]{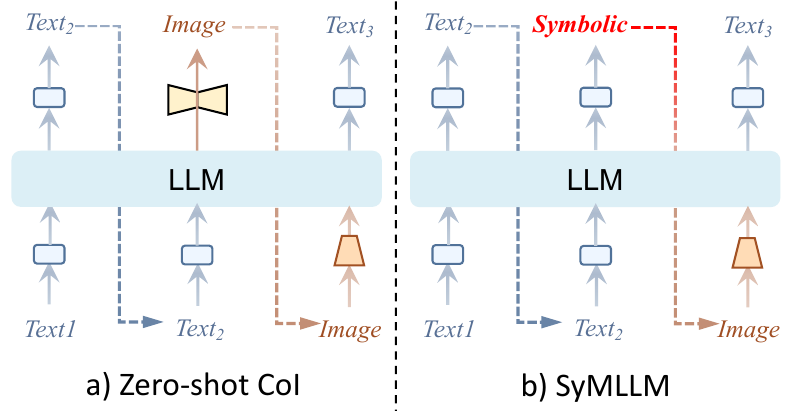}
    \caption{a) Given a uniform-modal model with a diffusion model as an image decoder, CoI can work under zero-shot mode. b) The symbolic response generated by SyMLLM given language instruction can be directly converted to an image.}
    \label{fig:symllm}
\end{minipage}
\end{figure}

However, the language-based reasoning process is often overly complex and abstract. Using images as a medium to solve textual logical problems is very intuitive. 
Humans excel at recalling visual concepts. For instance, one can easily recall someone's appearance while swiftly forgetting their name.
The human brain is inherently adept at intuitive reasoning.
We constantly employ imagery to facilitate thought processes in our daily lives. 
And many academic disciplines, as illustrated in Figure~\ref{fig:helpful_image}, utilize images to aid understanding and comprehension. 

Drawing inspiration from the human brain's proficiency in intuitive reasoning, we have developed the Chain-of-Images (CoI) method to generate images to help solve complex and abstract problems.
As illustrated in the right top of Figure~\ref{fig:cot-coi}, we generate images for the three shapes step by step. At the least two steps, we identify the intersection points between the newly generated shape and the shapes from previous steps in the image. While algorithmically computing intersection points is challenging for LLMs, with the images, discerning the count of intersection points is very easy for an image encoder with good pattern recognition capabilities.
As a preliminary attempt, we point out that our Chain of Images method broadly offers at least but far more than these three advantages:
\begin{itemize}
    \item Images provide an intuitive representation of the relationships (spatial, topological, temporal, etc.) between the items for tasks requiring complex relational reasoning, such as counting the intersection points of geometric shapes.
    \item Images assist in providing a compact representation that can record the current state at each step for tasks that satisfy Markovian properties and involve extended reasoning chains, like playing chess.
    \item Images capture a wealth of commonsense knowledge not present in language alone. Incorporating images into the reasoning process enhances its robustness.
\end{itemize}

Unfortunately, there currently aren't any publicly available unified multimodal models capable of understanding and generating images that meet the quality requirements.
In supporting our CoI reasoning, we introduce a symbolic multimodal large language model (SyMLLM) shown in Figure~\ref{fig:symllm} that generates images strictly based on language instructions and accepts both text and image as input.

Furthermore, to validate the value of CoI, we have developed a CoI evaluation dataset, including 15 distinct tasks. This dataset is intended to quantitatively assess the extent of the images generated by large multimodal models can facilitate reasoning, as opposed to merely evaluating image quality through human visual inspection.

Empirical results on geometric, chess, and commonsense tasks from the CoI evaluation dataset demonstrate that the SyMLLM framework can generate the required images with nearly 100\% % accuracy.
Based on such accurate image generation, the SyMLLM employing CoI outperforms the same structured LLM that relies solely on pure text for reasoning on these datasets.
For instance, the accuracy of calculating the number of intersection points among four shapes increased from 27.75\% to 64.25\%.

\section{Chain of Images Evaluation Dataset}
Current MLLMs~\citep{wu2023next,dong2023dreamllm,zheng2023minigpt5} have made significant progress in generating high-quality image details and understanding images. 
Yet, in many tasks, a graphical imagination is as important as or even more important to humans as an intermediate step in reasoning than a sequential one. 
The next frontier for these models is evaluating and enhancing their capacity for integrated reasoning with text and images.
To evaluate this capability, we introduced the CoI evaluation dataset, or CoIEval, which measures explicitly MLLMs' proficiency in utilizing image generation as a step in textual reasoning processes.

\begin{figure}[hbt]
    \includegraphics[width=\textwidth]{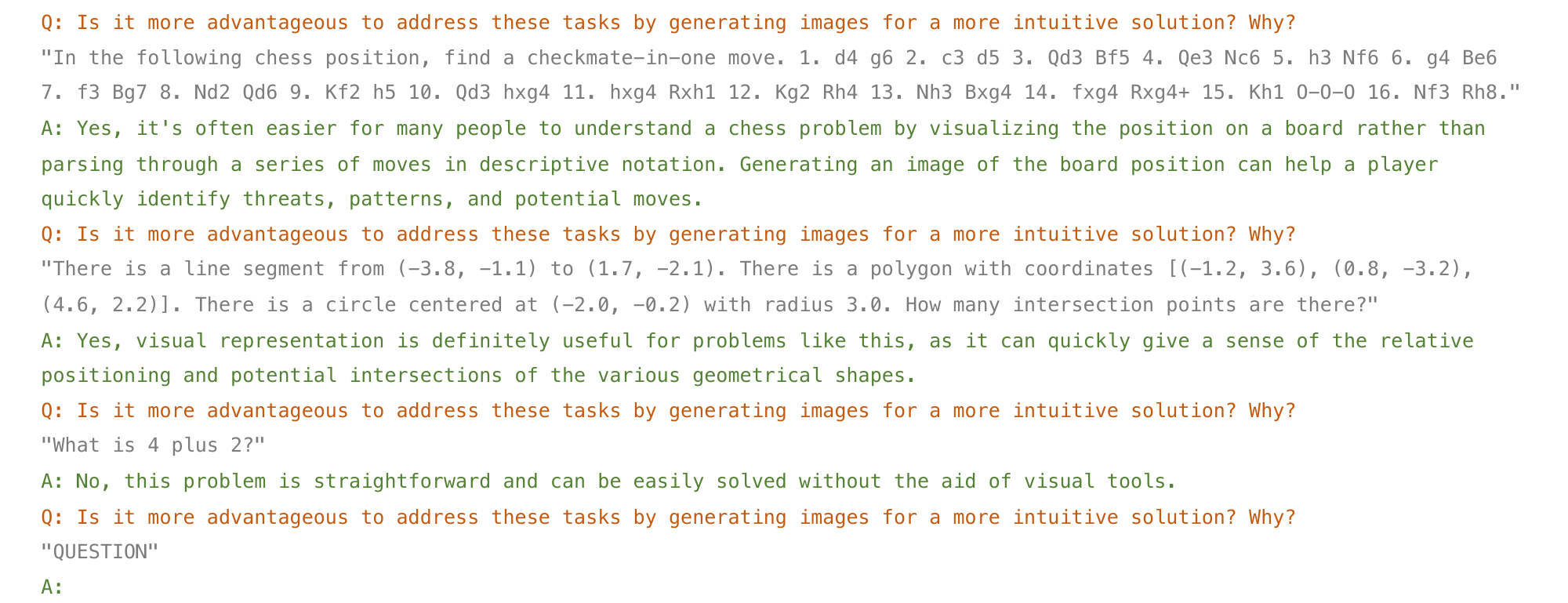}
    \caption{the 3-shot prompt for building the CoIEval dataset.}
    \label{fig:prompt_evalset}
\end{figure}

We employ GPT-4 to filter through various existing evaluation datasets, including BIGBench~\citep{srivastava2022beyond}, AGIEval~\citep{zhong2023agieval}, MMLU~\citep{hendrycks2020measuring}, %OpenBookQA~\citep{mihaylov2018can}, CommonSenseQA~\citep{talmor2018commonsenseqa}, 
NYCC\cite{hessel2023androids} to isolate questions suitable for CoI tasks. 
To guide GPT-4 in recognizing when image generation can be beneficial for problem-solving, we designed a specialized 3-shot prompt. 
This prompt contains three distinct questions, depicted in Figure \ref{fig:prompt_evalset}. 
It serves as an instruction for GPT-4 to identify where visual aids can enhance understanding and problem resolution. 
For example, problems in chess and geometry demonstrate how image generation can facilitate the comprehension of complex spatial relationships, whereas a simple arithmetic problem exemplifies a case where such visual support is unnecessary.
The "QUESTION" should be replaced by the questions from each task.

Based on the 3-shot prompt, we get a collection of 15 tasks across various domains, as shown in Appendix \ref{tab:coi_eval_set}. In this table, each row is divided into three lines:
Line 1. The task name and the source dataset from which the task is derived.
Line 2. An example question from the task.
Line 3. GPT-4's response regarding whether image generation aids in reasoning—indicating 'Yes' or 'No', along with the rationale for why GPT-4 believes an image would or would not be helpful.

\section{Symbolic Multimodal Large Language Models}
\begin{figure}[thbp]
  \centering
  \begin{subfigure}{0.45\textwidth}
    \centering
    \includegraphics[width=0.32\linewidth]{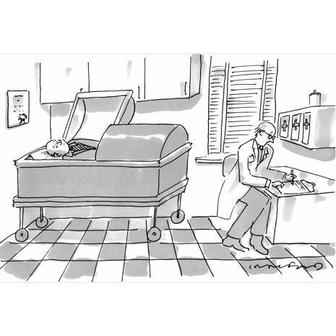}
    \includegraphics[width=0.32\linewidth]{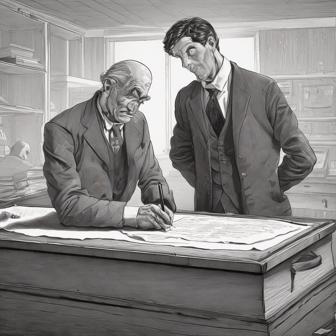}
    \includegraphics[width=0.32\linewidth]{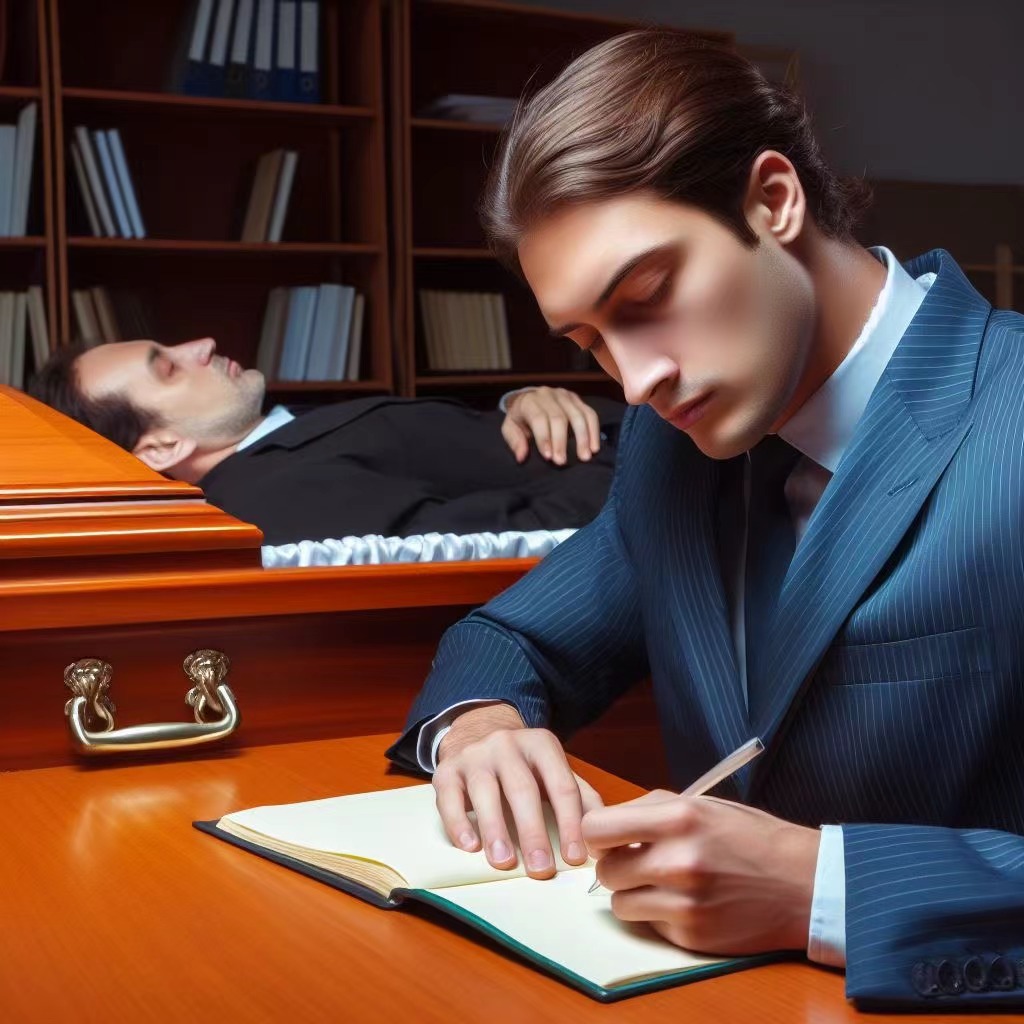}
    \caption{``A man at a desk writing something down. Another man is \textcolor{red}{lying behind} him \textcolor{red}{in} a coffin.''
    ``SDXL fail at processing the location relation between two men and a coffin.''}
    \label{subfig:coffin}
  \end{subfigure}%
  \hspace{0.05\linewidth} 
  \begin{subfigure}{0.45\textwidth}
    \centering
    \includegraphics[width=0.32\linewidth]{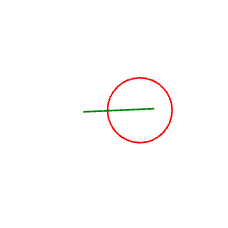}
    \includegraphics[width=0.32\linewidth]{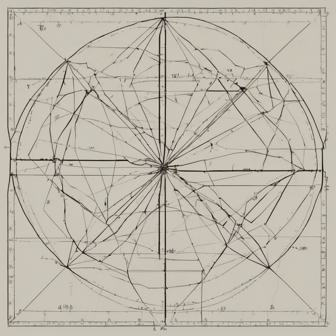}
    \includegraphics[width=0.32\linewidth]{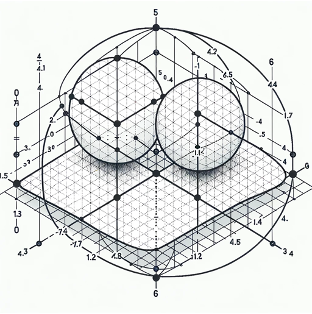}
    \caption{There is a \textcolor{red}{circle} centered at (1.6, -0.2) with radius 2.0. There is a \textcolor{red}{line segment} from (2.5, -0.3) to (-1.9, -0.1). How many intersection points are there?
    ``SDXL and DALL·E 3 unable to generate simple geometric shapes.''}
    \label{subfig:circle}
  \end{subfigure}
  \caption{This figure demonstrates the generation of new images by SDXL (center) and DALL·E 3 (right), using the captions derived from the original images (left).}
  \label{fig:sd_failure_case}
\end{figure}

The primary step of CoI is to generate images step by step.
Stable Diffusion XL~\citep{podell2023sdxl} are state-of-the-art open-source image generative models which good at creating images with vast details. 
However, as illustrated in Figure~\ref{subfig:coffin}, we encountered a notable limitation of the SDXL model when it was tasked with following complex instructions to generate images. 
Additionally, in Figure~\ref{subfig:circle}, SDXL could not draw simple shapes like a circle and a line segment.
The first limitation could potentially be addressed by DALL·E 3, an advanced close-source image generation model from OpenAI. However, even DALL·E 3 encounters difficulties when generating abstract sketches or symbolic diagrams that require adherence to strict rules or relational constraints, as also evidenced in Figure~\ref{subfig:circle}. 
Without an appropriate image, CoI cannot enhance the model's reasoning capabilities because introducing noisy and disturbing images could, on the contrary, worsen the performance. Therefore, we call for an accurate and controllable image generation strategy. Based on this request, we introduce the Symbolic Multimodal LLM (SyMLLM), which is shown in Figure~\ref{fig:symllm}. 
It consists of an LLM, a symbol-to-image decoder, and an image encoder. 
When various text prompts are provided to the LLM, it will produce symbolic representations in different formats (e.g., SVG format), which can be directly transformed into bitmap images. 
Subsequently, these images are converted into image embeddings by the image encoder. The embeddings are then concatenated with the text embeddings to predict the next token.
SyMLLM leverages the robust capabilities of LLMs to generate symbols precisely from language context and losslessly converts these symbols into image formats. In subsequent experiments, we observe that the accuracy of image generation approaches nearly 100\%. %Such a level of precision is inconceivable when using MLLM in conjunction with a diffusion model.

The symbol we use in this paper is Scalable Vector Graphics (SVG).
SVG is a widely adopted XML-based format for creating two-dimensional graphics. 
Unlike bitmap graphics which are made up of pixels, the raw file of SVG graphics are composed of language that can directly generated by LLMs.
Meanwhile, they can be shown as pixel images scaled up or down without any loss in quality. 
SVG offers the capability to produce graphics that range from simple shapes and icons (e.g. geometric shapes)to highly intricate and detailed illustrations (e.g. chess board and pieces).
With the help of the SVG format, we can generate images only using the text-based LLM.

\begin{figure}[hbt]
    \includegraphics[width=\textwidth]{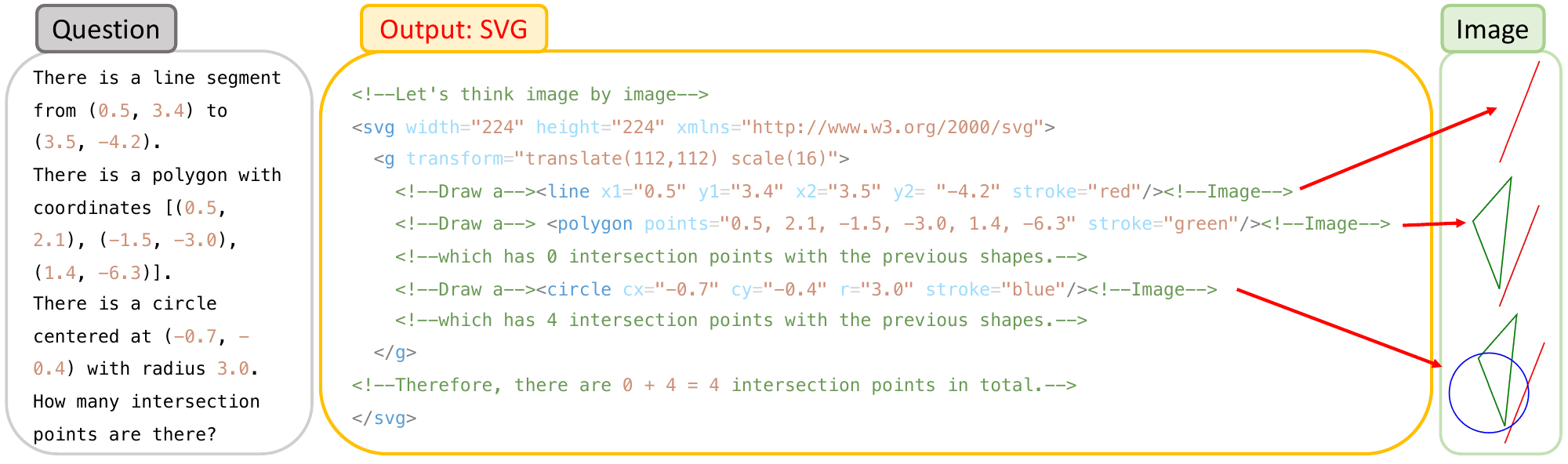}
    \caption{SVG format basic geometric shapes.}
    \label{fig:svg_prompt}
\end{figure}

As shown in Figure~\ref{fig:svg_prompt}, for the problem of counting the intersections of geometric shapes, the output of SyMLLM is a vector image in SVG format. 
The instructions are written as code comment in the vector image: $<!\!-\!-Let's~think~by~image -\!-\!>, <!\!-\!-Draw~a-\!-\!>, <!\!-\!-Image-\!-\!>, <!\!-\!-which~has~4~intersection~points~with~the~previous~shapes.-\!-\!>, <!\!-\!-Therefore,~there~are~0 + 4 = 4~intersection~points~in~total.-\!-\!>$. 
These instructions can be generated by the large model while do not affect the content of the pixel image converted. 
When come to $<!\!-\!-Image-\!-\!>$ tokens, we convert the vector images into pixel images and then input them into SyMLLM's image encoder. 
Based on the images, it is easy to compute the intersection points of the geometric shapes.

\begin{figure}[hbt]
    \includegraphics[width=\textwidth]{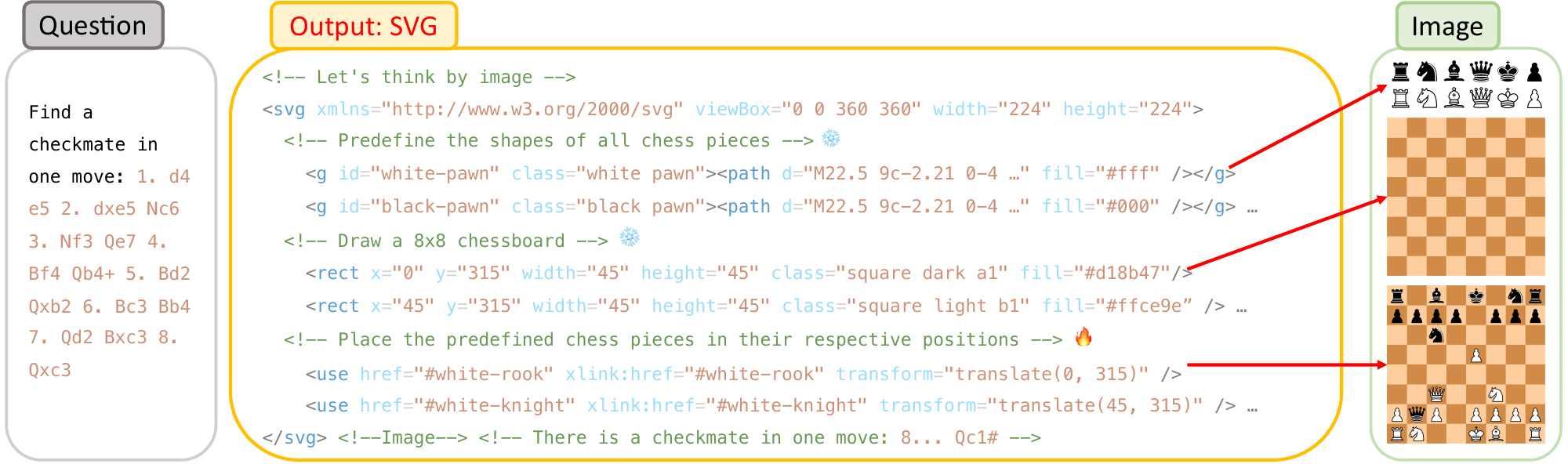}
    \caption{SVG format chess board and pieces.}
    \label{fig:chess_svg_prompt}
\end{figure}

As shown in Figure \ref{fig:chess_svg_prompt}, for the problem of predicting checkmate in one move, SVG first defines classes for different chess pieces using polygons, then draws an 8$\times$8 chessboard, and finally moves each piece to its final position. 
Since the first two steps are fixed for any chess position, we can provide them as prompts to the large model and only need to generate the final position of each piece. 
When $<!\!-\!-Image-\!-\!>$ tokens appear, we convert the current state of the chessboard into a pixel image, which is then input into SyMLLM's image encoder. 
Based on the current state, it is easy to predict whether a checkmate is possible in the next move.

\section{Experiments}
In this section, we have chosen three specific tasks from our CoIEval for evaluating the benefits of the CoI approach: Geometry, Chess, and Commensense reasoning tasks. 
\subsection{Geometric}

\textbf{Task Description}:
We conducted a comprehensive evaluation of the CoI approach to determine its efficacy in solving geometric problems. 
The evaluation was based on the Intersect Geometric task from the CoIEval. 
This task is designed to assess the ability of models to identify intersections among geometric shapes like line segments, circles and polygons. 
It contains a total of 250,000 examples, which are stratified into five levels of difficulty. The complexity is determined by the number of geometric shapes involved in each problem, which ranges from 2 to 6.

\textbf{Data Processing}:
For the purpose of evaluation, we constructed a balanced set by randomly selecting 1,000 examples from each level of difficulty to create an evaluation set. 
Additionally, we generated a training set by selecting 10,000 examples from each difficulty level, ensuring that there was no overlap between the sets due to the method of non-replacement sampling. 
In total, the training set comprised 50,000 examples and was converted into a format outlined in Figure~\ref{fig:svg_prompt}, whereas the evaluation set remained in its original form.

\textbf{Model Structure}:
For our experiments, we utilized the Vicuna-7B-v1.5~\citep{zheng2023judging}, an enhanced derivative of Llama 2~\citep{touvron2023llama}, pre-trained on user-shared conversations collected from ShareGPT\footnote{https://sharegpt.com}. The image encoder is the clip-vit-large-patch14~\citep{radford2021learning}. To establish a baseline for comparison, we trained a text-only model with identical hyper-parameters and prompts, excluding images during both training and inference.

\textbf{Training Details}:
We following~\citet{dettmers2023qlora} applied 4-bit quantization to the LLM's backbone and used a rank-16 LoRA, focusing on training newly added parameters. The fine-tuning process encompassed both the ViT and LLM, utilizing a single A800 GPU, with a batch size of 2 and 8 gradient accumulation steps, spanning over 5 epochs and completed within 12 hours.

\textbf{Experiment Results}:
The results are listed in Table~\ref{tab:ig-eval}. From the table, we can see that the SyMLLM framework achieves nearly 100\% accuracy in converting coordinates to an image. 
Moreover, the framework's ability to count intersection points within these images is significantly more effective compared to the baseline that relies solely on text. 
Specifically, the accuracy rate for detecting 4 intersection points has improved to 2.3$\times$, while the detection of 5 intersection points shows an improvement of 2.05$\times$.

\begin{table}[htb]
    \centering
    \caption{The intersect geometric task. The first row lists the number of geometric shapes in each sample, representing the different difficulty levels; The second and third rows compare the performance w/ or w/o image training. The bottom row reports the similarity between the image generated by SyMLLM and the ground truth.}
    \begin{tabular}{cccccc}
    \toprule
    Number of Geometry Shapes &2&3&4&5&6\\
    \bottomrule
    Text Accurcy&90.75&48.25&27.75&23.25&16.0\\
    CoI Accurcy&\textbf{95.5}&\textbf{85.75}&\textbf{64.25}&\textbf{47.75}&\textbf{33.25}\\    
    \midrule
    Image Similarity&100&100&99.99&99.99&99.99\\
    \bottomrule
    \end{tabular}
    \label{tab:ig-eval}
\end{table}

\begin{figure}[hbt]
\begin{subfigure}{0.19\textwidth}
    \includegraphics[width=\textwidth]{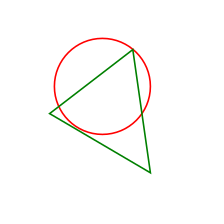}
    \caption{$2\to 4$}
\end{subfigure}
\begin{subfigure}{0.19\textwidth}
    \includegraphics[width=\textwidth]{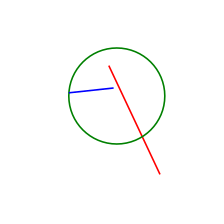}
        \caption{$2\to 1$}
\end{subfigure}
\begin{subfigure}{0.19\textwidth}
    \includegraphics[width=\textwidth]{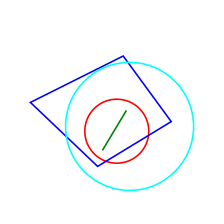}
        \caption{$4\to 6$}
\end{subfigure}
\begin{subfigure}{0.19\textwidth}
    \includegraphics[width=\textwidth]{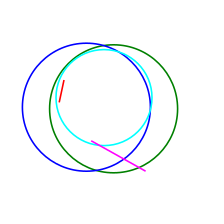}
        \caption{$7\to 9$}
\end{subfigure}
\begin{subfigure}{0.19\textwidth}
    \includegraphics[width=\textwidth]{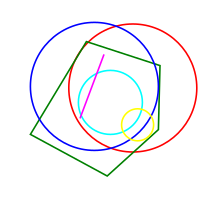}
        \caption{$14\to 18$}
\end{subfigure}
    \caption{Failure cases for 2 to 6 geometric shapes, respectively. The number preceding the arrow indicates the ground truth number of intersection points, while the number following the arrow shows the count by the SyMLLM.}
    \label{fig:failure_case}
\end{figure}

\textbf{Results Analysis}:
We have randomly selected one failure case from each level of difficulty and displayed it in Figure~\ref{fig:failure_case}. It is evident that these failure cases are challenging to discern, even for the human eyes.
One advantage of SVG lies in its scalability; it can be enlarged indefinitely without loss of quality. 
It is reasonable to believe that converting these graphics into larger images could further improve the performance of the CoI method on geometric intersection problems.

\begin{comment}
\begin{figure}[hbt]
\centering
    \includegraphics[width=\textwidth]{imgs/geo_split.pdf}
    \caption{The first row illustrates the question, answer and image. Prompting for geometric data (SVG format).
    And the second row illustrates the question, answer and image. Prompting for Chess data (FEN format)
    }
    \label{fig:symbolic_prompt}
\end{figure}
\end{comment}
\subsection{Chess}
\textbf{Task Description}:
We assess the effectiveness of CoI on the ``Checkmate in One Move" task from the CoIEval benchmark. 
This task challenges models to identify a single move within a chess position that would result in a checkmate. 
In the context of chess, ``checkmate" signifies a condition where the king is under immediate threat of capture with no legal moves available to escape. 
This task's complexity stems from the intricate series of moves leading up to the checkmate, which can involve a comprehensive logical sequence.

\textbf{Data Processing}:
We have categorized the data from the validation set into groups based on every 10 moves, with ``one move" indicating that each player has made a single move. 
For instance, a state with 33 moves would fall into the [31, 40) moves category.
Given the limited size of the dataset, it is solely utilized as a validation set. 
We derived our training dataset from a subset of data used in the research by ~\citet{feng2023chessgpt} on ChessGPT. 
This subset includes games in Portable Game Notation (PGN) format from the Pro-player dataset\footnote{https://www.pgnmentor.com} and the CCRL dataset\footnote{https://ccrl.chessdom.com}. 
We meticulously filtered this data to exclude games with non-standard openings, any illegal moves, or those not culminating in checkmate. 
The final counts of questions in the training and test sets are displayed in the last two rows of Table~\ref{tab:chess-eval}.
The final moves, which represent checkmates, are labeled as answers, and the preceding moves as questions. 
This training set was then converted into SVG format as depicted in Figure~\ref{fig:chess_svg_prompt}.

\textbf{Model Structure}:
Our SyMLLM is an extension of ChessGPT~\citep{feng2023chessgpt}, which was fine-tuned on GPT-NeoX-3B~\citep{black2022gpt} using a vast dataset related to chess games and language. 
SyMLLM incorporates an image encoder based on the clip-vit-large-patch14 model. 
The architecture is designed to first generate a symbolic representation in SVG format from the training set, then convert this SVG into an image. 
This image is subsequently processed by the image encoder, which aids in predicting the checkmate in one move based on the current state of the chessboard.

\textbf{Training Details}:
We following~\citet{dettmers2023qlora} applied 4-bit quantization to the LLM's backbone and used a rank-16 LoRA, focusing on training newly added parameters. 
The fine-tuning process encompassed both the ViT and LLM, utilizing a single A800 GPU, with a batch size of 1 and 8 gradient accumulation steps, spanning over 5 epochs and completed within 8 hours.

\textbf{Experiment Results}:
The performance of the SyMLLM on the "Checkmate in One Move" task is summarized in Table~\ref{tab:chess-eval}, showcasing the CoI's high proficiency. 
The table indicates that SyMLLM achieves an accuracy rate near 100\% in generating the correct chessboard state images. 
This accuracy facilitates the use of CoI reasoning to identify checkmate positions directly from the images, bypassing the need for complex textual reasoning chains. 
The CoI method significantly outperforms the baseline that relies on text-based inputs. 
For instance, the accuracy of [11,20) moves has improved to 1.38$\times$, and the accuracy of [61,70) moves has increased to 1.85$\times$. 

\begin{table}[htb]
    \centering
    \caption{Results on Checkmate in One Move dataset.}
    \begin{tabular}{ccccccccccc}
    \toprule
    Moves &1-10&11-20&21-30&31-40&41-50&51-60&61-70&71-80&81-90&91-100  \\
    \midrule
    Text Acc&19.71&29.03&26.3&28.47&35.15&33.01&36.94&50.0&50.0&33.33\\
    CoI Acc&\textbf{41.73}&\textbf{39.94}&\textbf{32.9}&\textbf{40.28}&\textbf{40.36}&\textbf{53.59}&\textbf{68.47}&\textbf{83.33}&\textbf{75.0}&\textbf{66.66}\\    
    \midrule
    Img Sim&100&99.93&99.63&99.67&99.93&99.92&99.93&99.9&99.91&99.92\\
    \midrule
    n\_train&1815&2200&3392&4428&3616&3043&2930&2618&1752&1134\\
    n\_test&137&651&1076&864&384&209&111&42&16&6\\
    \bottomrule
    \end{tabular}
    \label{tab:chess-eval}
\end{table}

\textbf{Results Analysis}:
When the representation of the chessboard state is consistently accurate, the length of the reasoning chain becomes irrelevant. 
As the number of moves in a sequence increases, the pool of legal moves narrows, enhancing the likelihood of successfully identifying a checkmate in one move. 
This is corroborated by the results in the table, where an increase in the number of steps correlates with improved outcomes.

\subsection{Commonsense}

\textbf{Task Description}:
Previous experiments have demonstrated that the Context of Inference (CoI) significantly aids in solving complex logical problems. 
In this section, we proceed to investigate whether images can enhance the capability of text in common sense reasoning. 
To this end, we have selected the Location and Unusual tasks for testing. 
These tasks utilized the same set of 531 event descriptions, but the questions differed. 
The Location task requires the description of an event, followed by a question about where the event took place. 
A model's answer is deemed correct only if it includes the provided standard answer. 
On the other hand, the Unusual task describes a scenario and then asks which parts of the description violate common sense. 
Given that questions about violations of common sense are open-ended, the accuracy of the answers cannot be directly calculated. 
Therefore, the study adopted different metrics: if the large-scale model fails to identify the nonsensical elements in the description, the answer is considered incorrect; otherwise, it is correct.

\textbf{Model Structure}:
While the SyMLLM is adept at accurately rendering abstract graphics, the generative model creates richly detailed, realistic images.
Therefore, we can use SDXL and DALL·E 3 to create common sense images without training a new model.
We use LLaVA-13B, a model obtained by finetuning Vicuna-13B and ViT on a text-image dataset, to recognize these common sense images and answer questions. 
The pure text model we compare against is Vicuna-13B. 
The CoI method employs the prompt, ``Let's solve the problem with the help of an image," to cue the model to use imagery for reasoning.

\textbf{Experiment Results}:
As shown in the following table: the use of pictures as a medium can introduce visual common sense information to help improve common sense reasoning problems. For example, using SDXL to generate pictures can improve the accuracy of determining the location of an event by 4.15\%. Using the more effective DALL-E 3 to generate pictures improves the accuracy by 9.66\%. Using the plain text model, 9.6\% of the samples could not find the anti-commonsense points in the textual descriptions, but it is easy to find them after using the pictures.

\begin{table}[htb]
    \centering
    \begin{tabular}{ccc}
        \toprule
         & Location Accuracy(\%)& Unusual Accuracy(\%)\\
         \midrule
         Text&73.63 &90.4\\
         CoI with SDXL&77.78 &\textbf{100}\\
         CoI with DALL·E 3&\textbf{82.29} &\textbf{100}\\
         \bottomrule
    \end{tabular}
    \caption{Caption}
    \label{tab:my_label}
\end{table}

\begin{figure}[hbt]
    \includegraphics[width=\textwidth]{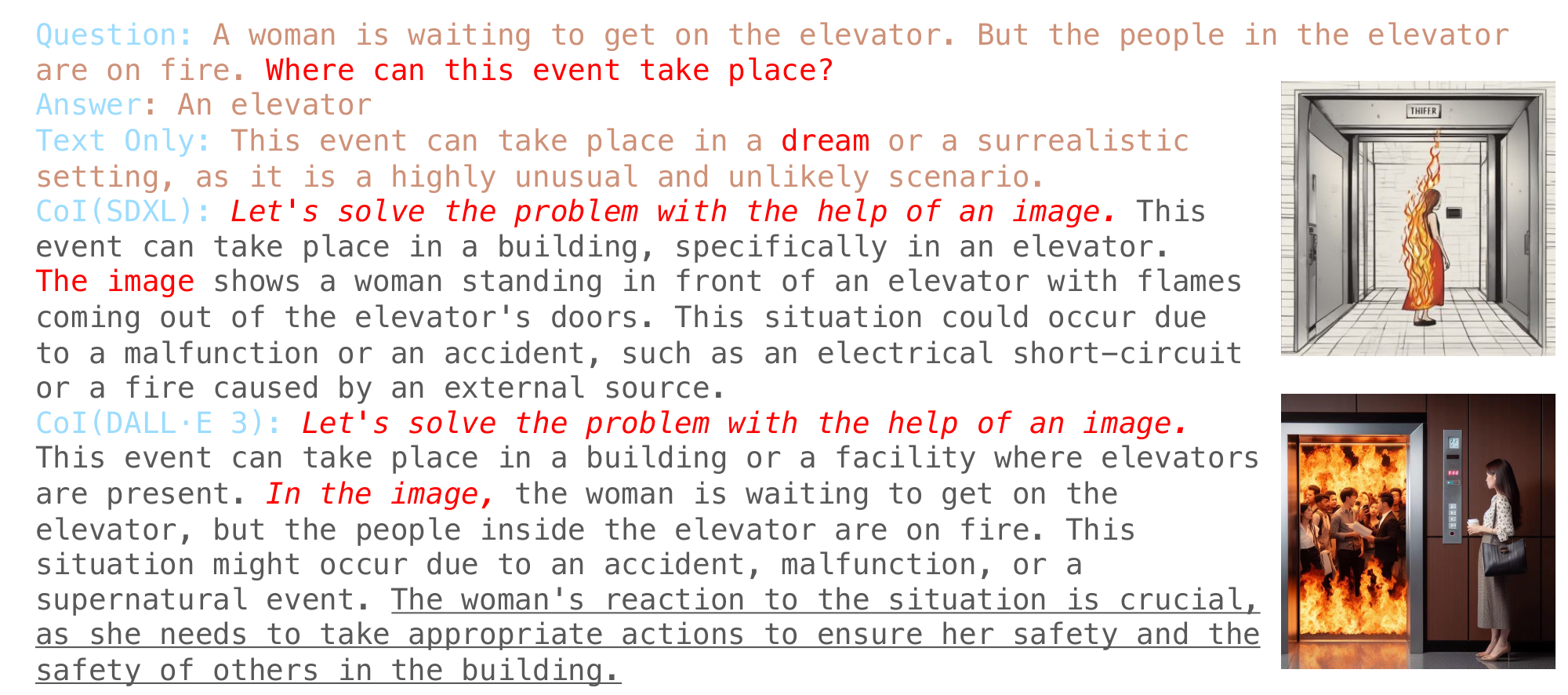}
    \caption{This is an example for the same question, the response of Text only reasoner, Chain of images with original picture and CoI with image generated by stable diffusion. CoI improves the performance of this problem}
    \label{fig:location}
\end{figure}

\begin{figure}[hbt]
    \includegraphics[width=\textwidth]{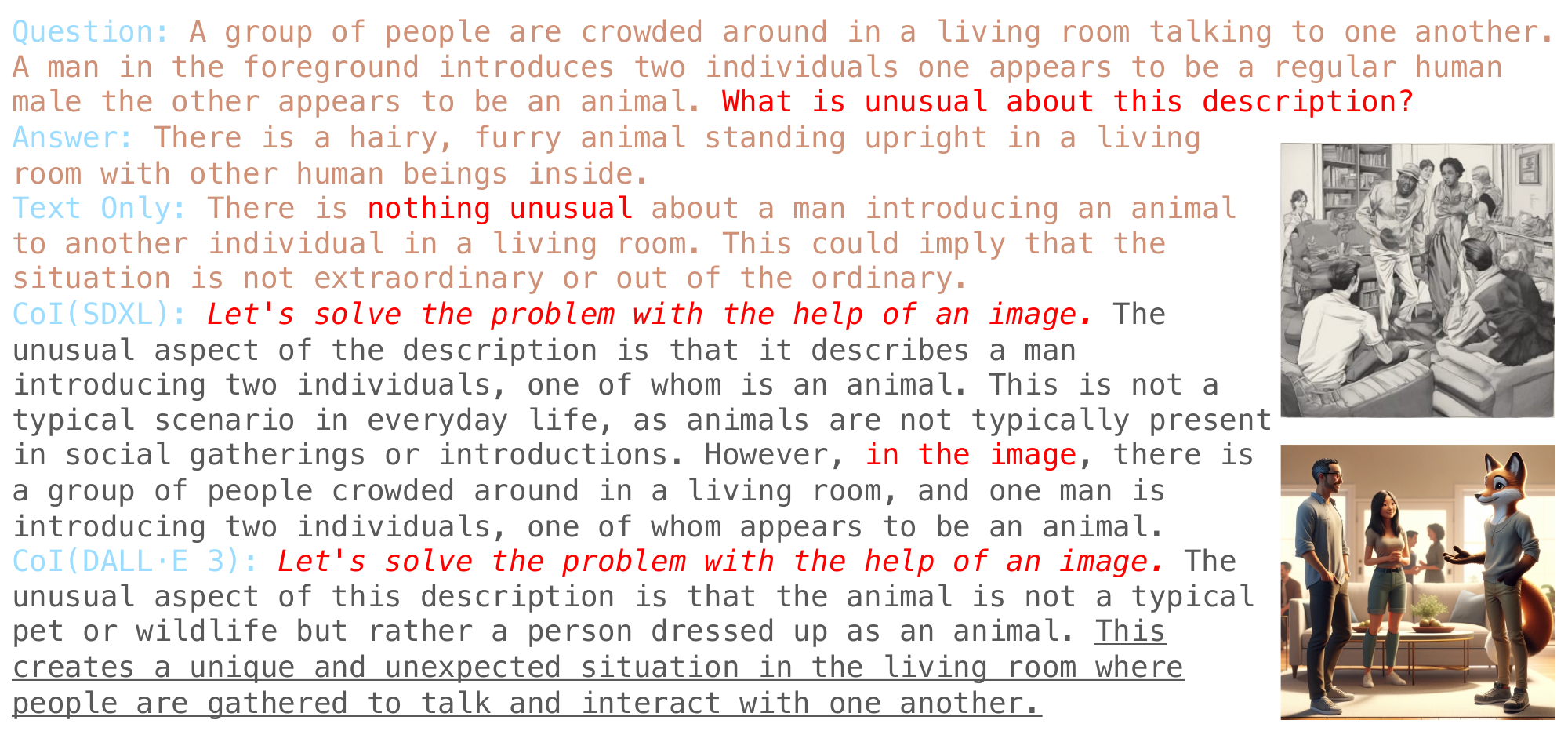}
    \caption{This is an example for the same question, the response of Text only reasoner, Chain of images with original picture and CoI with image generated by stable diffusion. CoI improve the performance of this problem}
    \label{fig:unusual}
\end{figure}

\textbf{Results Analysis}:
As shown in Figure \ref{fig:location} and Figure \ref{fig:unusual}, when solving the same problem, CoI is more intuitive compared to text-only reasoning and is able to supplement the missing details in the text with visual commonsense knowledge to aid the reasoning process. In addition, the strength of the image generation capability directly affects the reasoning efficiency of the CoI; for example, the CoI with DALL-E 3, which generates images of higher quality, is 4.51\% more accurate in recognizing the location of an event than the CoI with SDXL.

\section{Related Work}
\subsection{Reasoning with LLM}
\label{ssec:reasoning_with_llm}
With the emergence of LLMs with surprisingly powerful capabilities, it has become possible to use language models to address various non-traditional natural language tasks, such as some reasoning tasks: solving arithmetic problems, deducting about the commonsense, playing a game~\citep{openai2023gpt4, srivastava2022beyond, qin2023chatgpt}, etc. This ability is first manifested in the series of GPT model~\citep{radford2019language, brown2020language}, where the models address various tasks without fine-tuning by following the instruction of a paragraph of natural language called ``\textit{prompts}''. Centered around this prompt-leading generation capability, a series of works have been proposed to enhance the reasoning ability of large models using different forms of prompts: \citet{wei2022chain} introduced a chain-of-thought prompt which provides the model with learning examples of the target reasoning task containing the detailed reasoning steps, enabling the model to imitate the examples to generate answers to solve reasoning tasks—this capability is referred to as ``\textit{in-context learning}'' capability. Further, simply based on tasks described in natural language by the model (rather than examples), it can also stimulate the model's reasoning ability—this is called a 'zero-shot prompt' \citep{kojima2022large}. 

At the same time, some non-natural language prompting methods have been proposed. ``code prompts'' \citep{gao2023pal, hu2023code} are typical work, using code to help the model achieve better performance on some symbolic reasoning tasks. \citet{liu2022minds} further uses the code to operate simulation for some physical situations. \citet{nye2021work} use the intermediate state as the prompt to help the model address arithmetic problems. \citet{chen-2023-large} use a table as the prompt.
The idea behind this kind of work is that, compared to natural language, a more structured prompt can be more helpful for tasks with typical workflows. Unlike these works, we have noticed the limitations of using language and other sequential information to represent problems. We hope to point out that images such as flows, diagrams and illustrations might be another important type of information representation. For some tasks, these images can encode the necessary intermediate or step-by-step information for reasoning with greater information density and in a more appropriate manner.

\subsection{Multimodel large language model}
The monumental advancements in large language models within the domain of natural language processing have spurred a surge in efforts to adapt these successes to other modalities. Unlike these language models, multimodal models are designed to accept or produce input in various modalities beyond just natural language. These models incorporate additional modules to understand and generate content in formats such as images, voice, or video.

The underlying principle of this work involves integrating the pre-trained language model with another pre-train model for a specific modality, such as images. For instance, \citet{li2023blip} proposes to train a lightweight Transformer to align the image representation from an image encoder with the textual representation from an LLM. Drawing inspiration from the instruction tuning techniques used in pure language models, \citet{liu2023visual} employs data generated by GPT-4 to enhance question-answering performance. \citet{alayrac2022flamingo} achieves closer integration of image and text data by incorporating image information into the LLM through cross attention layer instead of embedding. Moreover, \citet{girdhar2023imagebind} proposes a joint embedding across six different modalities, including images, text, audio, depth, terminal, and IMU data, without the need for training on the last four modalities. This is achieved by leveraging the zero-shot capacity of the large image-text models. The above-mentioned models do not possess image-generation capabilities. To make up for this shortcoming, \citet{dong2023dreamllm} realize image generation by combining the model proposed by \citet{liu2023visual} with the Stable diffusion~\citep{Rombach_2022_high}. Building on this, \citet{wu2023next} further integrates it with the model from~\citet{girdhar2023imagebind}, utilizing its ability to align multiple modalities to generate and accept multiple modalities.
Very recently, GPT-4~\citep{openai2023gpt4} has demonstrated its astounding ability to understand and generate in the modalities of images and voice. In the examples it showcased, the GPT-4 has already reached a level close to human capabilities, adding both confidence and pressure to the research in this field.

Although our methodology is similar to these works, our motivation leans more towards using images as a tool to assist large models in reasoning (similar to CoT, code, and other descriptions of the intermediate process of reasoning we mentioned in Section~\ref{ssec:reasoning_with_llm}), rather than focusing on its multimodal interaction capabilities. Therefore, compared to the quality of generating text, images, and voice, we are more concerned about whether and in which tasks these inputs and outputs can better assist the model in completing its reasoning.

\section{Conclusions}
In this paper, we put forward a Chain-of-Image (CoI) prompting method, which imitates humans' ability to solve complex reasoning problems with the help of images.
To accurately generate images to help the CoI reasoning, we propose the SyMLLM framework. 
We perform experiments on three kinds of datasets: Geometric, Chess and Commonsense, using the SyMLLM framework to demonstrate the power of CoI.
We found that CoI uniformly improves the performance over pure-text models without images as an intermediate representation.
We believe our CoI method has great potential to help LLMs achieve human-level reasoning ability.

\newpage
\newpage

\bibliography{iclr2024_conference}
\bibliographystyle{iclr2024_conference}

\newpage

\appendix
\label{app}
\begin{table}[htbp]
    \centering
    \tiny
    \begin{tabular}{p{\textwidth}}
        \toprule
        Checkmate in One @ BIGBench\\
        Example: In the following chess position, find a checkmate-in-one move. 1. f4 Nc6 2. Nf3 Nf6 3. d4 d5 4. e3 Bd7 5. Be2 e6  \\
        GPT-4 Response: Yes, generating an image of the board position can help a player to quickly identify threats, patterns, and potential moves.\\
        \midrule
        Chess State Tracking @ BIGBench\\ 
        Example: For each of the following chess games, please complete the notation for the last shown move by filling in the destination square: d2d4 e7e6 e2e3 c7c5 f2f4 b7b6 g1f3 d7d6 f1e2 a7a6 e1g1 g7g6 c2c3 f8g7 c1d2 g8e7 d2e1 e8g8 e3e4 c8b7 b1d2 f7f5 d1b3 b7 \\ 
        GPT-4 Response: Yes, generating an image of the chess board would allow for a more intuitive understanding of the game's progression and the positions of the pieces.\\
        \midrule
        Geometric Shapes @ BIGBench\\ 
        Example: This SVG path element \textless path d="M 31.94,78.63 L 66.91,49.50 L 68.54,41.07 L 61.03,39.02 M 61.03,39.02 L 52.78,44.98 M 52.78,44.98 L 31.94,78.63"/\textgreater  draws a [sector, heptagon, octagon, circle, rectangle, hexagon, triangle, line, kite, pentagon]  \\
        GPT-4 Response: Yes, an image would provide a clear, visual representation of the shape, making it easier to identify.\\
        \midrule
        Intersect Geometry @ BIGBench\\ 
        Example: Find the number of intersection points between the shapes and lines specified by the coordinates given. There is a line segment from (-0.1, -2.2) to (-2.4, -2.4). There is a circle centered at (1.8, 1.5) with radius 2.0. How many intersection points are there? \\
        GPT-4 Response: Yes, by plotting the line and the circle on a graph, one can easily see where they intersect, especially for those who are more visually inclined.\\
        \midrule
        Logic Grid Puzzle @ BIGBench\\ 
        Example: There are 2 houses next to each other, numbered 1 on the left and 2 on the right. There is one person living in each house. The people in these houses have different characteristics: Each person plays a different sport: one is a soccer player and one is a baseball player. Each person has a different device: one has a radio and one has a gameboy. Clues: 1. The baseball player lives in the first house. 2. The person who has a gameboy does not live in the first house. What is the number of the house where the person who has a gameboy lives?  \\
        GPT-4 Response: Yes, it can help to visualize the positions of the houses and the characteristics of the people living in them, making it easier to solve the problem.\\
        \midrule
        Logical Deduction @ BIGBench\\ 
        Example: The following paragraphs each describe a set of five objects arranged in a fixed order. The statements are logically consistent within each paragraph.
        On a shelf, there are five books: a green book, a blue book, a white book, a purple book, and a yellow book. The blue book is to the right of the yellow book. The white book is to the left of the yellow book. The blue book is the second from the right. The purple book is the second from the left.  \\
        GPT-4 Response: Yes, visualizing the arrangement of the books can make it easier to understand their relative positions and to answer questions about the order.\\
        \midrule
        Matrix Shapes @ BIGBench\\ 
        Example: Compute the kronecker product of a matrix of shape (2,2,3,3) with a matrix of shape (2,2,4,2). Add the result to a matrix of shape (4,4,12,6). Sum the result over the second axis. Compute the kronecker product of the result with a matrix of shape (2,3,3). \\
        GPT-4 Response: Yes, generating images can help in visualizing the shapes and dimensions of the matrices involved, making it easier to comprehend and solve the problem.\\
        \midrule
        Movie Dialog Same or Different @ BIGBench\\ 
        Example: The following is a conversation between two people, but the transcript doesn't mark who said what:
        "You're asking me out." "That's so cute." "What's your name again?" "Forget it." ----- In the preceding conversation, were the sentences "That's so cute." and "What's your name again?" said by the same or different individuals? \\
        GPT-4 Response: Yes, it can help to visually organize the conversation and make it easier to track who might be speaking at any given time.\\
        \midrule
        Reasoning About Colored Objects @ BIGBench\\
        Example: On the table, you see a bunch of objects arranged in a row: a grey cat toy, a brown paperclip, an orange envelope, a gold notebook, a red teddy bear, and a magenta mug. What is the color of the object directly to the right of the gold object?\\
        GPT-4 Response: Yes, it allows for a more intuitive understanding of the spatial arrangement of the objects, making it easier to identify the object to the right of the gold object.\\
        \midrule
        AQuA-RAT @ AGIEval\\
        Example: A car is being driven, in a straight line and at a uniform speed, towards the base of a vertical tower. The top of the tower is observed from the car and, in the process, it takes 10 minutes for the angle of elevation to change from 45° to 60°. After how much more time will this car reach the base of the tower? \\
        GPT-4 Response: Yes, visual representation can be helpful for this problem. Drawing a diagram with the tower, car's positions, and angles of elevation can make it easier to apply trigonometric concepts and visualize the distances and relationships between the car and tower.\\
        \midrule
        LogiQA-EN @ AGIEval\\
        Example: In the planning of a new district in a township, it was decided to build a special community in the southeast, northwest, centered on the citizen park.These four communities are designated as cultural area, leisure area, commercial area and administrative service area.It is known that the administrative service area is southwest of the cultural area, and the cultural area is southeast of the leisure area. Based on the above statement, which of the following can be derived?\\
        GPT-4 Response: Yes, in cases like this where spatial relationships are described, a visual representation can be very helpful. By mapping out the relative positions of the communities, it can provide a clearer understanding of the arrangement and can assist in deriving accurate conclusions based on the provided information.\\
        \midrule
        LSAT-AR @ AGIEval\\
        Example: Of the eight students—George, Helen, Irving, Kyle, Lenore, Nina, Olivia, and Robert—in a seminar, exactly six will give individual oral reports during three consecutive days—Monday, Tuesday, and Wednesday. Exactly two reports will be given each day—one in the morning and one in the afternoon—according to the following conditions: Tuesday is the only day on which George can give a report. Neither Olivia nor Robert can give an afternoon report. If Nina gives a report, then on the next day Helen and Irving must both give reports, unless Nina's report is given on Wednesday.", "question": "Which one of the following could be the schedule of the students' reports?"\\
        GPT-4 Response: Yes, for this kind of logic-based puzzle, visual representation like a chart or a table can be beneficial. It allows for easier organization of information and constraints, making the process of elimination and arrangement more intuitive.\\
        \midrule
        High School Mathematics @ MMLU\\
        Example: If a pentagon P with vertices at (– 2, – 4), (– 4, 1), (–1, 4), (2, 4), and (3, 0) is reflected across the line y = x to get a new pentagon, P', then one of the vertices of P’ is\\
        GPT-4 Response: Yes, by plotting the points and reflecting them across the line, it becomes easier to identify the new position of the vertices and determine which one corresponds to P'.\\
        \midrule
        Location @ NYCC\\
        Example: In a tree, a mother bird is tending to her nest and babies. There is another bird in a cage on an upper branch. Where can this event take place?\\
        GPT-4 Response: Yes, the key of this problem is that the location should allow for trees where birds can naturally make nests and also accommodate a caged bird. An image could help illustrate this scenario, making it more tangible and perhaps easier to understand, especially for visual learners.\\
        \midrule
        Unusual @ NYCC\\
        Example: A group of people are crowded around in a living room talking to one another. A man in the foreground introduces two individuals one appears to be a regular human male the other appears to be an animal. What is unusual about this description?\\
        GPT-4 Response: Yes, an image can add context and may help highlight the oddity of an animal being introduced as if it were a person, which is the unusual aspect of this description.\\
        \bottomrule
    \end{tabular}
    \caption{Datasets that GPT-4 believes generate images would be helpful.}
    \label{tab:coi_eval_set}
\end{table}

\end{document}